\documentclass[lettersize,journal]{IEEEtran}
\usepackage{CJKutf8}
\usepackage{amsmath,amsfonts}
\usepackage{algorithmic}
\usepackage{algorithm}
\usepackage{array}
\usepackage[caption=false,font=normalsize,labelfont=sf,textfont=sf]{subfig}
\usepackage{textcomp}
\usepackage{stfloats}
\usepackage{url}
\usepackage{verbatim}
\usepackage{graphicx}
\usepackage{tabularx}
\usepackage{cite}
\usepackage{multirow}
\usepackage{booktabs}
\usepackage{mathrsfs}  
\usepackage{makecell}
\usepackage{color}
\usepackage[numbers, sort & compress]{natbib} 

\begin{document}


\title{\LARGE \bf
Visual SLAMMOT Considering Multiple Motion Models}

\author{Peilin~Tian$^{2}$ and Hao~Li$^{*1,2}$ 
	\thanks{$^{1}$Department of Automation, Shanghai Jiao Tong University (SJTU), Shanghai, 200240, China.}
	\thanks{$^{2}$\'Ecole d’Ing\'enieurs Paris SJTU (SPEIT), Shanghai, 200240, China.}
	\thanks{$^{*}$Corresponding author: Hao Li (email: haoli@sjtu.edu.cn).}
}

\maketitle

\begin{abstract}
Simultaneous Localization and Mapping (SLAM) and Multi-Object Tracking (MOT) are pivotal tasks in the realm of autonomous driving, attracting considerable research attention. While SLAM endeavors to generate real-time maps and determine the vehicle's pose in unfamiliar settings, MOT focuses on the real-time identification and tracking of multiple dynamic objects. Despite their importance, the prevalent approach treats SLAM and MOT as independent modules within an autonomous vehicle system, leading to inherent limitations. Classical SLAM methodologies often rely on a static environment assumption, suitable for indoor rather than dynamic outdoor scenarios. Conversely, conventional MOT techniques typically rely on the vehicle's known state, constraining the accuracy of object state estimations based on this prior. To address these challenges, previous efforts introduced the unified SLAMMOT paradigm, yet primarily focused on simplistic motion patterns. In our team's previous work IMM-SLAMMOT\cite{IMM-SLAMMOT}, we present a novel methodology incorporating consideration of multiple motion models into SLAMMOT i.e. tightly coupled SLAM and MOT, demonstrating its efficacy in LiDAR-based systems. This paper studies feasibility and advantages of instantiating this methodology as visual SLAMMOT, bridging the gap between LiDAR and vision-based sensing mechanisms. Specifically, we propose a solution of visual SLAMMOT considering multiple motion models and validate the inherent advantages of IMM-SLAMMOT in the visual domain.
\end{abstract}


\section{Introduction}
\label{sec:intro}

\begin{figure}[!t]
\centering
\subfloat[]{\includegraphics[width=0.45\textwidth]{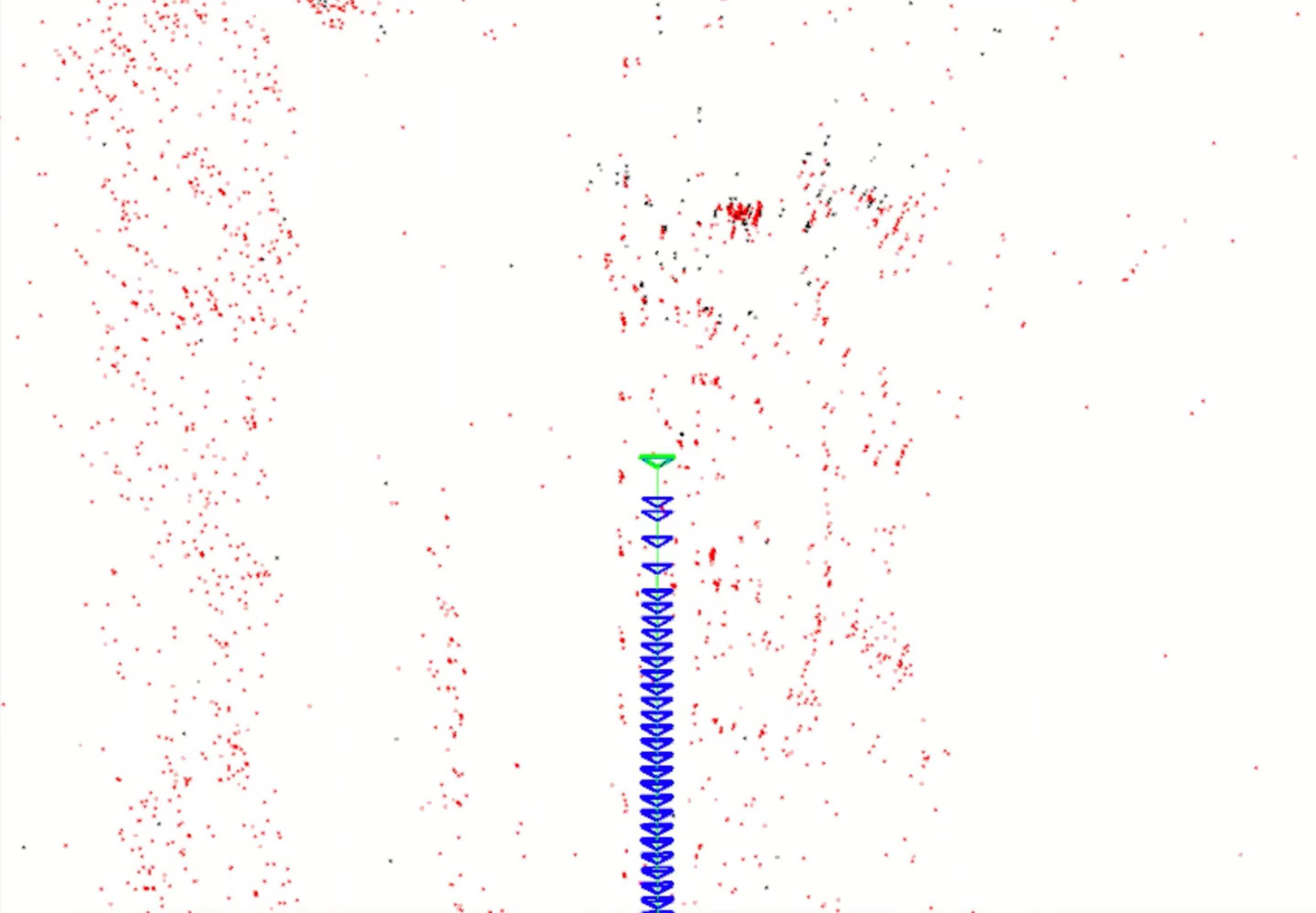}%
\label{0018-map}}
\hfil
\subfloat[]{\includegraphics[width=0.45\textwidth]{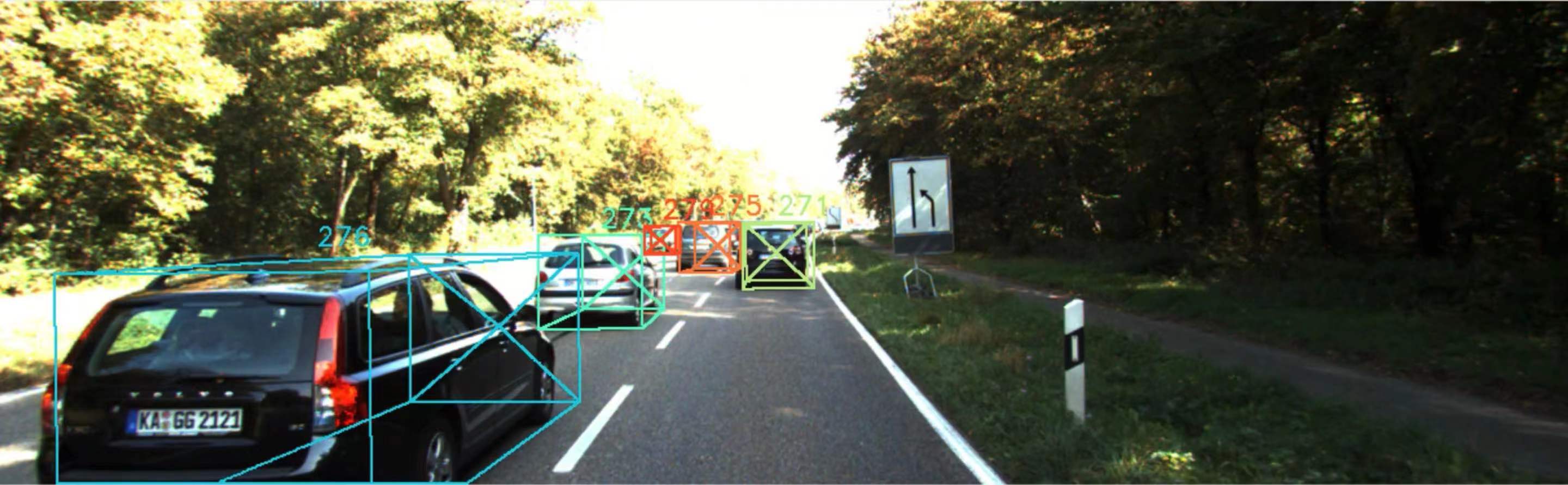}%
\label{0018-mot}}
\caption{Output of the proposed framework, containing the camera trajectory and a static environmental map (a), as well as the 3D object states (b).}
\label{0018-slammot}
\end{figure}

\IEEEPARstart{S}{imultaneous} Localization and Mapping (SLAM) and Multi-Object Tracking (MOT) are two essential tasks in autonomous driving, garnering significant attention in recent years. SLAM aims to construct real-time maps and estimate the vehicle's pose in unknown environments, while MOT focuses on identifying and tracking multiple dynamic objects in real-time. These tasks are both crucial for achieving highly autonomous and intelligent vehicle systems.

Currently, SLAM and MOT are generally operated as two independent modules within an intelligent vehicle system. However, this paradigm imposes certain limitations on both of these tasks. On the one hand, classical SLAM frameworks are often based on the static environment assumption, which is suitable for indoor rather than outdoor scenarios. For autonomous driving, moving surrounding vehicles can interfere with the vehicle's localization and introduce unnecessary motion information into the constructed environmental map. On the other hand, prevalent MOT research typically treats the vehicle's state as known information and estimates object states based on this prior. In this case, the estimated object states are constrained by the accuracy of the vehicle's state. Therefore, in real-world driving scenarios, SLAM and MOT, as two modules requiring real-time operation, are intrinsically linked and mutually dependent.

Some studies have attempted to implement SLAM and MOT as a unified framework and proposed the more general paradigm of SLAMMOT\cite{wang2007simultaneous}. SLAM and MOT can respectively be regarded as reduced cases of SLAMMOT. This paradigm is more suitable for the common outdoor dynamic environments in autonomous driving and can enhance the performance of both SLAM and MOT simultaneously. However, previous methods often consider simple object motion patterns, such as a single constant velocity model. In realistic environments, environmental objects often exhibit complex motion states and switch between different patterns. At this point, simple motion models are insufficient to effectively describe the object's state. Therefore, in our team's previous work\cite{IMM-SLAMMOT}, we propose a methodology coined as Methodology Level 3 that consists in incorporating consideration of multiple motion models into SLAMMOT i.e. tightly coupled SLAM and MOT.

It is worth noting that SLAMMOT can be implemented based on various sensors. Common sensors include cameras and LiDARs. Many LiDAR-based SLAM and MOT methods can achieve good results. However, visual sensors inherently lack depth information, making tasks in 3D environments more challenging. Besides, intelligent vehicle systems based on different sensors exhibit significant differences at the algorithmic level. Our team’s previous work\cite{IMM-SLAMMOT} presents instantiation of the Methodology Level 3 as LiDAR SLAMMOT and demonstrates advantages of this methodology for LiDAR-based systems. Since LiDAR sensors and visual sensors are essentially different and the two kinds of sensor systems have completely different sensing mechanisms and properties, whether instantiating the Methodology Level 3 as visual SLAMMOT is indeed feasible and, if so, whether the Methodology Level 3 is indeed advantageous for visual SLAMMOT are two questions of which the answers are neither known nor evident.

This paper answers the questions, as its two contributions. First, this paper provides a solution of SLAMMOT taking advantage of the Methodology Level 3 to demonstrate feasibility of instantiating this methodology as visual SLAMMOT. Second, this paper demonstrates that advantages of the Methodology Level 3 indeed exist for visual SLAMMOT.

\section{Related Work}
\label{sec:relatedwork}

\subsection{ Visual SLAM }
\label{subsec:vslam}

Visual SLAM methods typically consist of modules such as sensor data acquisition, frontend odometry, backend optimization, mapping, and loop closure. Common sensors include monocular cameras, stereo cameras, and RGB-D cameras. 

Different visual SLAM methods vary in the frontend visual odometry techniques. Some methods are based on feature points, requiring keypoint extraction and descriptor computation for each frame to match features between adjacent frames, facilitating subsequent camera localization and map construction. Typical methods of this paradigm include MonoSLAM\cite{MonoSLAM}, PTAM\cite{PTAM}, and the ORB-SLAM series\cite{ORB-SLAM-1,ORB-SLAM-2,ORB-SLAM-3}. Among them, ORB-SLAM is based on ORB features, utilizes bundle adjustment for optimization, and features a comprehensive keyframe selection strategy. Moreover, ORB-SLAM incorporates a loop closure thread to enhance the global consistency of localization and mapping. Compared to previous feature-based visual SLAM methods, ORB-SLAM shows significant improvements in the SLAM task and is adaptable to various visual sensors as well as IMU. Some methods omit descriptor computation in visual odometry, relying instead on photometric consistency between pixel-level data in consecutive frames, and estimate the camera motion by minimizing photometric error. Such methods are referred to as direct or semi-direct methods, with representative works including DTAM\cite{DTAM}, LSD-SLAM\cite{LSD-SLAM}, SVO\cite{SVO}, and DSO\cite{DSO}.

Such visual SLAM methods are based on the static environment assumption. While this assumption holds true for indoor environments, it presents notable shortcomings in autonomous driving scenarios. Moving objects in the roadway environment can interfere with the localization and mapping of intelligent vehicles, which becomes a challenge that pure SLAM methods fail to effectively address.

SLAM methods (or pure SLAM methods) form Methodology Level 0 that will be presented below. A specific SLAM method can serve as a concrete instantiation of Methodology Level 0 itself and can also serve as a concrete composing module of Methodology Level 1, Methodology Level 2, and Methodology Level 3 that will also be presented below.

\subsection{Visual MOT}  \label{subsec:vmot}

MOT is crucial for applications in dynamic environments \cite{Vu2009} \cite{Darms2009} \cite{Aeberhard2012} \cite{Li2013d}. State-of-the-art MOT methods typically use deep learning models to estimate the 2D or 3D information of objects. 2D objects can be represented precisely by pixel-level masks, which is obtained using image segmentation models. 3D objects can be represented by 3D bounding boxes, which is achieved through 3D object detection and tracking models.

\textbf{2D Image Segmentation}. Image segmentation aims to partition different regions within a 2D image. In recent years, deep learning-based image segmentation methods facilitate the analysis of semantic information in different parts of images. Currently, the main sub-directions of image segmentation include semantic segmentation, instance segmentation and panoptic segmentation. Semantic segmentation aims to predict the semantic category for each pixel, with common solutions based on Convolutional Neural Networks (CNNs)\cite{FCN, DeepLabv2}. Semantic segmentation can divide different regions in images according to semantic categories yet cannot distinguish between different instances of the same category. To address this issue, instance segmentation combines semantic segmentation with object detection, aiming to finely label each object with pixel-level masks. Some methods follow a top-down paradigm\cite{DeepMask, Mask-R-CNN}, first obtaining object regions through detection mechanisms and then performing semantic segmentation within these regions. Other methods adopt a bottom-up paradigm, embedding each pixel and segmenting through clustering algorithms\cite{SGN, SpatialEmbedding}. Additionally, some models perform instance segmentation in a one-stage manner\cite{YOLACT1, SOLO1}, achieving higher efficiency. In recent years, the task of panoptic segmentation has gradually gained attention\cite{EfficientPS, Panoptic-DeepLab}. Compared to instance segmentation, panoptic segmentation further segments areas in the image that do not belong to countable objects, enriching the algorithm's understanding of image content.

\textbf{3D Object Detection and Tracking}. Visual sensors inherently lack depth information, making vision-based 3D perception tasks challenging. With the advancement of deep learning, numerous methods become able to infer 3D information of objects directly from 2D images. Firstly, 3D object detection necessitates acquiring information on the size, position, and orientation of objects in the environment. The configuration of different visual sensors determines the model's input, leading to various visual 3D object detection paradigms. Among these, models based on monocular cameras\cite{Mono3D, MonoDLE} face significant challenges in achieving 3D object detection, yet they are less demanding on hardware requirements and exhibit advantages in real-time performance. Stereo cameras aid in inferring depth information from images, and such models\cite{LIGA-Stereo, DSGN} can achieve higher detection accuracy. In recent years, Bird's-Eye View (BEV)-based multi-camera 3D object detection models\cite{DETR3D, BEVFormer} have garnered attention. These methods can detect objects around the vehicle, not just limited to the front, and leverage Transformer\cite{attention} to achieve higher accuracy. Nonetheless, such models impose higher hardware and computational demands, necessitating increased deployment costs. On the other hand, the objective of 3D multi-object tracking is to temporally associate objects based on the foundation of object detection, assigning consistent IDs to the same object across different frames. Some methods adopt a two-stage paradigm\cite{AB3DMOT, MoMA-M3T} by associating existing detection results. These tracking methods facilitate integration with advanced object detection models, yet the tracking performance is simultaneously constrained by the detection results. Another category of single-stage methods\cite{QD-3DT, MUTR3D} employs a unified network, treating detection and tracking jointly as a single task, to fully utilize the information from the state of surrounding objects.

MOT methods (including image segmentation methods) themselves do not form any of the four Methodology Levels, yet a MOT method or an image segmentation method can serve as a concrete composing module of Methodology Level 1, Methodology Level 2, and Methodology Level 3 that will be presented below.

\subsection{Visual SLAMMOT}

Bearing in mind limitations of the static environment assumption, researchers have attempted to integrate classical SLAM methods with MOT methods. Wang et al.\cite{wang2007simultaneous} first define the problem of Simultaneous Localization, Mapping, and Moving Object Tracking (SLAMMOT) and implement the framework using traditional algorithms. In recent years, extensive work has focused on combining visual SLAM with deep learning models to enhance localization and mapping in dynamic environments, and to provide richer scene understanding to intelligent vehicles.

Handling moving objects in dynamic environments is crucial for SLAMMOT methods. From the methodology perspective, existing SLAMMOT methods may be roughly categorized into four methodology levels\cite{IMM-SLAMMOT}:

\textbf{Methodology Level 0: No cooperation between SLAM and MOT} (namely to treat all objects as stationary in SLAM). Traditional SLAM methods belong to this category, which has been reviewed in Section \ref{subsec:vslam}.

\textbf{Methodology Level 1: Slight cooperation between SLAM and MOT} (namely to use certain detection or segmentation method to determine potential moving objects and then to heuristically sift out potential moving objects in SLAM). Typical works include\cite{Detect-SLAM, DS-SLAM, DynaSLAM}.
Detect-SLAM\cite{Detect-SLAM} utilizes an object detection module integrated into ORB-SLAM2\cite{ORB-SLAM-2}, and identifies moving objects through semantic information and probabilistic descriptions. DS-SLAM\cite{DS-SLAM} employs semantic segmentation within ORB-SLAM2\cite{ORB-SLAM-2}, identifying and eliminating moving objects by semantic information and moving consistency check. DynaSLAM\cite{DynaSLAM} incorporates an instance segmentation module into ORB-SLAM2\cite{ORB-SLAM-2} for prior filtering of dynamic objects during localization. The aforementioned methods filter out dynamic objects from the SLAM process, allowing multi-object tracking to be independently carried out based on the SLAM results.
Once SLAM is realized separately, then based on ego-vehicle poses revealed by the SLAM composing module, MOT can be realized using whatever method such as reviewed in Section II-B.

\textbf{Methodology Level 2: Holistic cooperation between SLAM and MOT considering a single motion model} (namely to holistically fuse state estimation of moving objects and SLAM while ﬁtting moving objects rigidly with a single motion model). Typical works include\cite{CubeSLAM, DynaSLAM-2, VDO-SLAM, MOTSLAM}.
CubeSLAM\cite{CubeSLAM} is based on ORB-SLAM2\cite{ORB-SLAM-2} and utilizes deep learning models to detect and track three-dimensional cube objects. This method leverages object representations and motion model constraints to jointly optimize the poses of the camera and objects along with environmental points. DynaSLAM II\cite{DynaSLAM-2} is an improved version of DynaSLAM\cite{DynaSLAM}, which achieves joint optimization of camera pose, dynamic object trajectories and static scene structure through a novel bundle adjustment method. VDO-SLAM\cite{VDO-SLAM} employs instance segmentation and dense optical flow estimation as preprocessing modules. To enhance the robustness of camera and object state estimation, this method employs optical flow for joint estimation. Similar to DynaSLAM II, VDO-SLAM utilizes graph optimization to tightly couple localization, mapping, and object tracking. MOTSLAM\cite{MOTSLAM} utilizes various deep learning methods including 2D detection, 3D detection, semantic segmentation and depth estimation. The obtained dynamic object information assists in feature association and object representation, while SLAM and MOT are coupled through bundle adjustment. The aforementioned methods adopt the same paradigm: deep learning models are used to identify potential moving objects, followed by joint optimization of SLAM and MOT. Typically, in these optimization frameworks, only a constant velocity model is used to describe the motion state of objects.

\textbf{Methodology Level 3: Holistic cooperation between SLAM and MOT considering multiple motion models} (namely to holistically fuse state estimation of moving objects and SLAM while ﬁtting moving objects ﬂexibly with multiple motion models). Considering multiple motion models, which enables SLAMMOT to effectively handle the inherent non-deterministic nature of moving objects, is the essential point of this methodology. Instantiating this methodology as LiDAR-based SLAMMOT is presented in \cite{IMM-SLAMMOT}, whereas this paper focuses on instantiating this methodology as visual SLAMMOT.

It is worth noting that Methodology Level 3 contains two logic points: First, holistic cooperation between SLAM and MOT (in contrast with ``separate'' or ``loosely-coupled'' SLAM and MOT as in Methodology Level 1) is important. In fact, Methodology Level 2 also contains this first logic point. Second, considering multiple motion models is important. The first logic point has long since been pointed out by pioneer researchers, yet it does not necessarily hold, if only considering a single motion model, especially in the context of visual SLAMMOT. It is combination of both the first logic point and the second logic point that demonstrates power and advantage over the other three Methodology Levels.

\section{Architecture Overview}
\label{sec:architecture}

\begin{figure*}[!t]
\centering
\includegraphics[width=\textwidth]{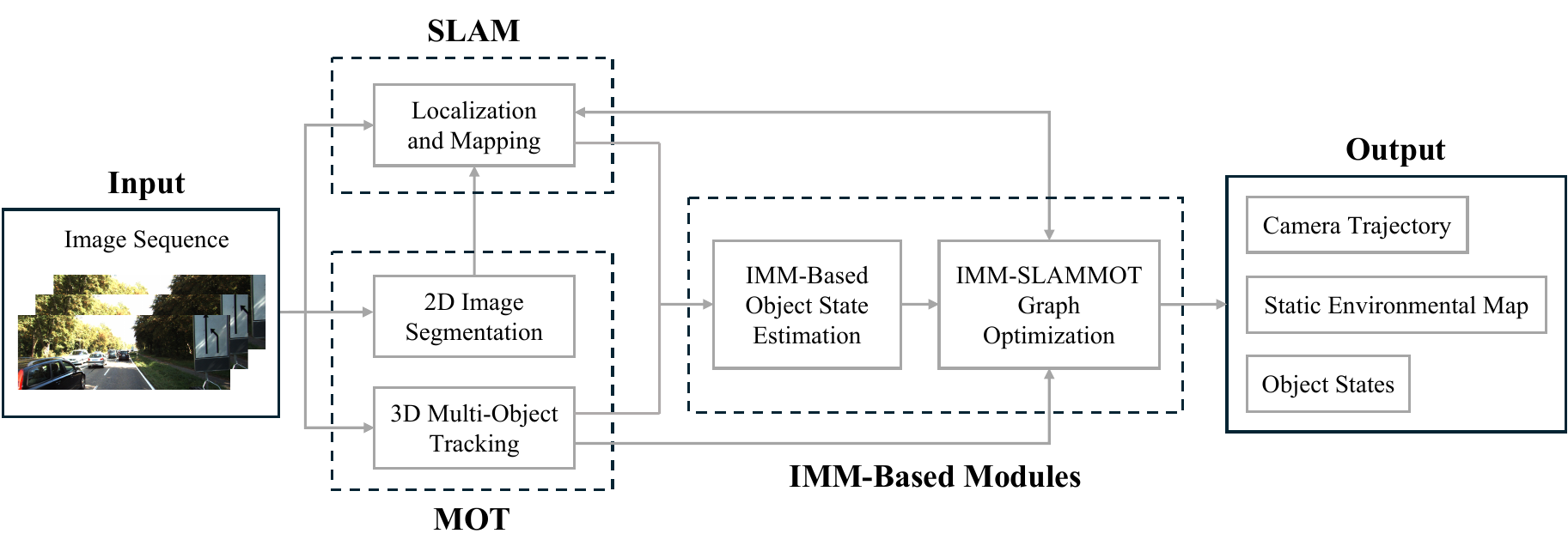}
\caption{Proposed framework, taking an image sequence as input to estimate camera trajectory, static environmental map and object states.}
\label{architecture}
\end{figure*}

In this section, we provide an overview of the methodology and its internal modules. We specifically focus on visual sensors in this paper, such as monocular cameras, stereo cameras and RGB-D cameras. Our method takes a sequence of images as input and then conducts online processing on these images through three procedures, including MOT,
SLAM and IMM-based modules. The output of our method corresponds to the SLAMMOT task, which contains the states of the camera and objects at different timestamps, as well as a global map of the static environment. The illustration of our framework is shown in Fig. \ref{architecture}.

\subsection{SLAM}

SLAM serves as a fundamental module in our method, implemented based on ORB-SLAM2\cite{ORB-SLAM-2}. This module is designed for online estimation of the camera trajectory and construction of the environmental map. Specifically, the SLAM module initializes the camera pose and environmental map at the beginning of the process. In each subsequent frame, ORB keypoints and descriptors are extracted and computed, then matched with the previously obtained ones. The matched features are used to estimate the camera poses and to maintain map points simultaneously in the world coordinate system. In ORB-SLAM2, local and global bundle adjustments are implemented using factor graph optimization. The nodes in the graph include camera poses and map points. In the following IMM-SLAMMOT graph optimization module, these factors are jointly optimized and updated with object states.

\subsection{MOT}

The purpose of the MOT module
is to provide prior information about objects for the following SLAMMOT task. This module consists of two parts: 2D image segmentation and 3D multi-object tracking. On the one hand, the goal of 2D segmentation is to obtain semantic information about objects in the images. We adopt the instance segmentation model SpatialEmbedding\cite{SpatialEmbedding} for implementation. This model takes a single 2D image as input to classify and cluster pixels in a bottom-up manner, generating pixel-level masks for each individual instance. As a result, we are able to determine the category of each pixel in the subsequent SLAM module. 3D multi-object tracking aims to provide prior object information, specifically the position, orientation, and bounding box of each object, while maintaining consistent object IDs between consecutive frames. To achieve this, we employ the two-stage MOT model MoMA-M3T\cite{MoMA-M3T}. In the first stage, 3D object detection results are obtained by MonoDLE\cite{MonoDLE}. Then, MoMA-M3T associates these detections in consecutive frames through motion modeling and learning. Output of the 3D MOT module is used in the subsequent IMM-based modules.

\subsection{IMM-Based Modules}

The IMM-based modules primarily consist of two parts. In the first object state estimation module, we take the position and the yaw rate of objects from the MOT
module, and transfer them from camera coordinates to world coordinates using the camera poses estimated by the SLAM module. Then, we employ an IMM-based filtering approach to estimate the motion states of these objects. The estimates are passed to the factor graph module, where they are jointly optimized with the camera poses and map points. The optimization results are utilized to update the relevant variables, aiming to improve both SLAM and MOT for subsequent frames. We will elaborate on the specific algorithms of the IMM-based modules in Section \ref{sec:imm}. It is worth noting that incorporating consideration of multiple motion models into SLAMMOT is the essential point of the Methodology Level 3. Concerning the IMM itself, like in \cite{IMM-SLAMMOT}, we just adopt it without scrutinizing any improved variant of it. The focus of this paper is to instantiate visual SLAMMOT considering multiple motion models, rather than achieving marginal improvements through other IMM variants.

\section{IMM-Based Object State Estimation \\ and Graph Optimization}
\label{sec:imm}

\subsection{Modeling}

To describe the states of objects, we follow our team's previous work \cite{IMM-SLAMMOT} to implement three motion models that are commonly observed in the real world: the Constant Position (CP) model, the Constant Velocity (CV) model and the Constant Turning Rate and Velocity (CTRV) model. We denote the set of motion models as $\mathcal{M} = \{\mathrm{CP}, \mathrm{CV}, \mathrm{CTRV}\}$. We utilize the coordinate system conventionally employed in visual SLAM, which follows the right-hand rule, with the $y$-axis pointing vertically downwards. Since the vertical motion during vehicle movement can be neglected, we only consider the horizontal motion in the $x-z$ plane. The object states of different models at timestamp $t$ are then defined as:
\begin{equation}
\left\{
\begin{aligned}
&\mathbf{x}^\mathrm{CP}_t = [x_t, z_t, \theta_t]^T \\
&\mathbf{x}^\mathrm{CV}_t = [x_t, z_t, \theta_t, v_t]^T \\
&\mathbf{x}^\mathrm{CTRV}_t = [x_t, z_t, \theta_t, v_t, \omega_t]^T
\end{aligned}
\right.
\end{equation}
where $\theta_t$, $v_t$ and $\omega_t$ denote the yaw rate, the linear velocity and the turning rate of an object at timestamp $t$, respectively.

We denote the motion of an object under model $d$ between two timestamps by a function $g^d$:
\begin{equation}
\label{eq:x=g(x)}
\mathbf{x}^d_t = g^d(\mathbf{x}^d_{t-1})\ ,\ d \in \mathcal{M}
\end{equation}

By definition of the considered motion models, we have
\begin{align}
\label{eq:motionCP}
g^\mathrm{CP}(\mathbf{x}^\mathrm{CP}_t) & = \mathbf{x}^\mathrm{CP}_t \\
\label{eq:motionCV}
g^\mathrm{CV}(\mathbf{x}^\mathrm{CV}_t) & = \begin{bmatrix}
                                      x_t + v_t{\rm cos}(\theta_t)\Delta T \\ 
                                      z_t + v_t{\rm sin}(\theta_t)\Delta T \\ 
                                      \theta_t \\ 
                                      v_t 
                                     \end{bmatrix} \\
\label{eq:motionCTRV}
g^\mathrm{CTRV}(\mathbf{x}^\mathrm{CTRV}_t) & = \begin{bmatrix}
                                       x_t + v_t{\rm cos}(\theta_t + \frac{\omega_t \Delta T}{2})\Delta T \\ 
                                       z_t + v_t{\rm sin}(\theta_t + \frac{\omega_t \Delta T}{2})\Delta T \\ 
                                       \theta_t + \omega_t \Delta T \\ 
                                       v_t \\
                                       \omega_t
                                      \end{bmatrix}
\end{align}

To apply the filtering method, we approximate (\ref{eq:x=g(x)}) by first order and we have
\begin{equation}
\mathbf{x}^d_t \simeq \mathbf{A}^d_{t-1} \mathbf{x}^d_{t-1} \ ,\ \mathrm{with}\ \mathbf{A}^d_{t} = \frac{\partial g^d(\mathbf{x}^d_t)}{\partial \mathbf{x}^d_t} \ ,\ d \in \mathcal{M} \\
\end{equation}
Note that the CP model is linear, thus there is no actual need for approximation.

\subsection{IMM-Based Object State Estimation}

For an object in the environment, suppose a timestamp $t-1$ for initialization of IMM. 

Each recursive cycle of the algorithm starts with three model weights $w^c_{t-1} = 1/3$, three state means $\hat{\mathbf{x}}^c_{t-1} = \mathbf{0}$ and three associated covariances $\hat{\mathbf{P}}^c_{t-1}$ of large values, with $c \in \mathcal{M}$. 

The following procedure of IMM-based object state estimation consists of four steps\cite{Li2022FARET} \cite{FARET}:

\subsubsection{Multi-model merging}
The model weights, state means and covariances are merged respectively in a weighted average manner based on their values at the previous timestamp. Denote $C^{cd}$ the transition probability from the model $c \in \mathcal{M}$ to the model $d \in \mathcal{M}$. These merged variables can be calculated as
\begin{subequations}  \label{eq:multi_model_merging}
\begin{align}
w^{d,M}_t & = \sum_{c \in \mathcal{M}} C^{cd} w^c_{t-1} \\
\mathbf{x}^{d,M}_t & = \frac{1}{w^{d,M}_t} \sum_{c \in \mathcal{M}} \hat{\mathbf{x}}^c_{t-1} C^{cd} w^c_{t-1} \\
\mathbf{P}^{d,M}_t & = \frac{1}{w^{d,M}_t} \sum_{c \in \mathcal{M}} C^{cd} w^c_{t-1} [\hat{\mathbf{P}}^c_{t-1} + \Delta \mathbf{x}^2_{t, cd}]
\end{align}
\end{subequations}
where
\begin{align*}
\Delta \mathbf{x}^2_{t, cd} = (\hat{\mathbf{x}}^c_{t-1} - \mathbf{x}^{d,M}_t)(\hat{\mathbf{x}}^c_{t-1} - \mathbf{x}^{d,M}_t)^T
\end{align*}

In practice, the transition probabilities $C^{cd}$ with $c,d \in \mathcal{M}$ can be represented as a matrix $\mathbf{C}$. The dimensions of the matrix $\mathbf{C}$ correspond to the number of models. Under consideration of three models, the matrix $\mathbf{C}$ can be defined as

\begin{equation}
\mathbf{C} = 
\begin{bmatrix}
1-2 \tau & \tau & \tau \\ 
\tau & 1-2 \tau & \tau \\ 
\tau & \tau & 1-2 \tau 
\end{bmatrix}
\end{equation}
where $\tau$ denotes the transition probability between two different models. In our implementation, the parameter $\tau$ is pre-set to 0.02.

\subsubsection{Distributed estimation}
For each model $d \in \mathcal{M}$, its previous estimate $\{\mathbf{x}^{d,M}_t, \mathbf{P}^{d,M}_t\}$ is used to predict its a priori estimate $\{\bar{\mathbf{x}}^d_t, \bar{\mathbf{P}}^d_t\}$ via the corresponding system model. The Extended Kalman Filter serves as the recursive estimation method for the three distributed estimation tracks\cite{Li2022FARET} \cite{FARET}.

\textbf{Prediction:}
\begin{subequations} \label{eq:EKF1}
\begin{align}
\bar{\mathbf{x}}^d_t & = \mathbf{A}^{d,M}_t \mathbf{x}^{d,M}_t \\
\bar{\mathbf{P}}^d_t & = \mathbf{A}^{d,M}_t \mathbf{P}^{d,M}_t (\mathbf{A}^{d,M}_t)^T + \mathbf{Q}^d
\end{align}
\end{subequations}
where the diagonal matrix $\mathbf{Q}^d$ denotes the system noise of the model $d$.

\textbf{Update:}
\begin{subequations}
\begin{align}
\hat{\mathbf{x}}^d_t & = \bar{\mathbf{x}}^d_t + \mathbf{K} (\mathbf{z}_t - \mathbf{H}^d \bar{\mathbf{x}}^d_t) \\
\hat{\mathbf{P}}^d_t & = (\mathbf{I} - \mathbf{K} \mathbf{H}^d) \bar{\mathbf{P}}^d_t
\end{align}
\end{subequations}
where $\mathbf{z}_t = [x^\mathbf{z}_t, z^\mathbf{z}_t, \theta^\mathbf{z}_t]^T$ denotes the measurement obtained from the MOT
module. $\mathbf{H}$ is the observation matrix, which is identity and maps from the dimension of $\mathbf{x}^d_t$ to the dimension of $\mathbf{z}_t$. $\mathbf{K}$ is the Kalman gain which is computed as
\begin{equation}  \label{eq:EKF5}
\mathbf{K} = \bar{\mathbf{P}}^d_t (\mathbf{H}^d)^T [\mathbf{H}^d \bar{\mathbf{P}}^d_t (\mathbf{H}^d)^T + \mathbf{R}]^{-1}
\end{equation}
where $\mathbf{R}$ is the measurement noise.

\subsubsection{Model weight update}
The weight $w^d_t$ is updated from its initial value $w^{d,M}_t$ according to its measurement innovation:
\begin{equation}
w^d_t = \eta \frac{w^{d,M}_t}{\sqrt{\mathrm{det} \mathbf{S}^d_t}} e^{-\frac{1}{2} [\mathbf{z}_t - \mathbf{H}^d \hat{\mathbf{x}}^d_t]^T (\mathbf{S}^d_t)^{-1} [\mathbf{z}_t - \mathbf{H}^d \hat{\mathbf{x}}^d_t]}
\end{equation}
where $\eta$ is a normalization constant and
\begin{equation}
\mathbf{S}^d_t = \mathbf{H}^d \hat{\mathbf{P}}^d_t (\mathbf{H}^d)^T + \mathbf{R}
\end{equation}

\subsubsection{Output synthesis}
The recursive cycle outputs the synthesized state estimate $\{\hat{\mathbf{x}}_t, \hat{\mathbf{P}}_t\}$ which is computed as
\begin{subequations}  \label{eq:output_synthesis}
\begin{align}
\hat{\mathbf{x}}_t & = \sum_{d \in \mathcal{M}} w^d_t \hat{\mathbf{x}}^d_t \\
\hat{\mathbf{P}}_t & = \sum_{d \in \mathcal{M}} w^d_t [\hat{\mathbf{P}}^d_t + (\hat{\mathbf{x}}^d_t - \hat{\mathbf{x}}_t) (\hat{\mathbf{x}}^d_t - \hat{\mathbf{x}}_t)^T]
\end{align}
\end{subequations}
Note that the state vector $\hat{\mathbf{x}}^d_t$ corresponding to different motion models has varying dimensions. Here, for clarity of expression, we assume that $\hat{\mathbf{x}}^d_t$ has the dimension of the most complex model ($\hat{\mathbf{x}}^d_t \in \mathbb{R}^5$). For state variables not included in the simpler models (i.e., $v_t$ and $\omega_t$), we set them to zero by default.

It is worth noting that estimate consistency \cite{Li2013a} has been considered implicitly in (\ref{eq:multi_model_merging} c) and (\ref{eq:output_synthesis} b).

\subsection{IMM-SLAMMOT Graph Optimization}

\begin{figure}[!ht]
\centering
\includegraphics[width=0.5\textwidth]{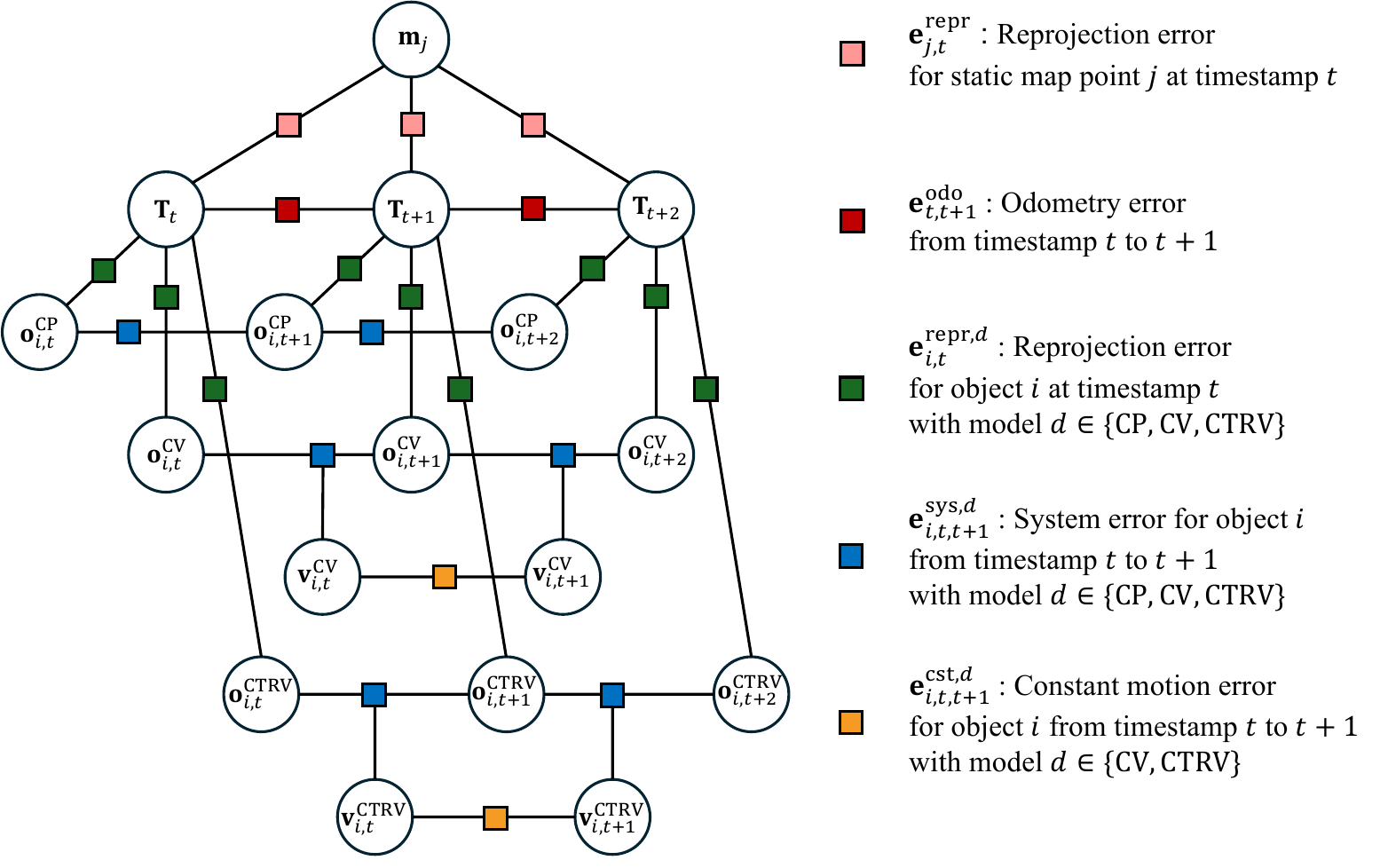}
\caption{IMM-SLAMMOT factor graph representation.}
\label{graph}
\end{figure}

To tightly couple SLAM and MOT for mutual benefits, we employ a novel bundle adjustment method based on graph optimization. We construct vertices for ego poses ($\mathbf{T} \in \mathbf{SE}(3)$), map points ($\mathbf{m} \in \mathbb{R}^3$), and object states ($\mathbf{o}$, $\mathbf{v}$) respectively, and connect different nodes through errors. To incorporate IMM, we set up states for each motion model and assign model weights to the corresponding edges. We include static map points as nodes in the factor graph, allowing the map to dynamically adjust with the optimization of ego poses, thereby improving the accuracy of subsequent localization and mapping. The structure of our proposed factor graph is illustrated in Fig. \ref{graph}.

Each static map point $\mathbf{m}_j \in \mathbb{R}^3$ can be reprojected to the image frames where it is observable. Denote $\mathbf{u}_{j,t}$ the coordinates of the pixel corresponding to this map point at timestamp $t$, and $\pi_t$ the reprojection function from a world point to a pixel. Then, the reprojection error of this map point can be calculated as
\begin{equation}
\mathbf{e}^{\mathrm{repr}}_{j,t} = \mathbf{u}_{j,t} - \pi_t(\mathbf{T}_t \bar{\mathbf{m}}_j)
\end{equation}
where $\bar{\mathbf{m}}_j \in \mathbb{R}^4$ represents the homogeneous coordinates of $\mathbf{m}_j \in \mathbb{R}^3$.

To take into account the initial visual odometry results, we define the odometry error between two consecutive timestamps as
\begin{equation}
\mathbf{e}^{\mathrm{odo}}_{t,t+1} = (\mathbf{T}^{t+1}_t \mathbf{T}_t)^{-1} \mathbf{T}_{t+1}
\end{equation}

For each motion model $d \in \mathcal{M}$, and for each object $i$ at timestamp $t$, we define its position ($\mathbf{p}^d_{i,t}$), its pose state ($\mathbf{o}^d_{i,t}$) and its measurement ($\mathbf{o}^\mathbf{z}_{i,t}$, obtained by MOT) as
\begin{subequations}
\begin{align}
\mathbf{p}^d_{i,t} & = [x^d_{i,t}, y^d_{i,t}, z^d_{i,t}] \in \mathbb{R}^3 \\
\mathbf{o}^d_{i,t} & = [x^d_{i,t}, y^d_{i,t}, z^d_{i,t}, \theta^d_{i,t}] = [\mathbf{p}^d_{i,t}, \theta^d_{i,t}] \in \mathbb{R}^4 \\
\mathbf{o}^\mathbf{z}_{i,t} & = [x^\mathbf{z}_{i,t}, y^\mathbf{z}_{i,t}, z^\mathbf{z}_{i,t}, \theta^\mathbf{z}_{i,t}] = [\mathbf{p}^\mathbf{z}_{i,t}, \theta^\mathbf{z}_{i,t}] \in \mathbb{R}^4
\end{align}
\end{subequations}

Then, the reprojection error for object $i$ with motion model $d$ can be calculated as
\begin{equation}
\mathbf{e}^{\mathrm{repr},d}_{i,t} = [[\mathbf{T}_t (\bar{\mathbf{p}}^d_{i,t})^T]^T_{1:3}, \theta^d_{i,t} - \phi_t]^T - (\mathbf{o}^\mathbf{z}_{i,t})^T
\end{equation}
where $\bar{\mathbf{p}}^d_{i,t} \in \mathbb{R}^4$ represents the homogeneous coordinates of $\mathbf{p}^d_{i,t} \in \mathbb{R}^3$, and $\phi_t$ denotes the yaw rate of the ego vehicle.

Besides, we define the full states $\mathbf{s}^d_{i,t}$ considering the velocities $\mathbf{v}^d_{i,t}$ for different motion models:
\begin{equation}
\mathbf{s}^\mathrm{CP}_{i,t} = \mathbf{o}^\mathrm{CP}_{i,t}
\end{equation}
\begin{equation}
\left\{
\begin{aligned}
&\mathbf{v}^\mathrm{CV}_{i,t} = v^\mathrm{CV}_{i,t} \\
&\mathbf{s}^\mathrm{CV}_{i,t} = [\mathbf{o}^\mathrm{CV}_{i,t}, \mathbf{v}^\mathrm{CV}_{i,t}] = [\mathbf{o}^\mathrm{CV}_{i,t}, v^\mathrm{CV}_{i,t}]
\end{aligned}
\right.
\end{equation}
\begin{equation}
\left\{
\begin{aligned}
&\mathbf{v}^\mathrm{CTRV}_{i,t} = [v^\mathrm{CTRV}_{i,t}, \omega^\mathrm{CTRV}_{i,t}] \\
&\mathbf{s}^\mathrm{CTRV}_{i,t} = [\mathbf{o}^\mathrm{CTRV}_{i,t}, \mathbf{v}^\mathrm{CTRV}_{i,t}] = [\mathbf{o}^\mathrm{CTRV}_{i,t}, v^\mathrm{CTRV}_{i,t}, \omega^\mathrm{CTRV}_{i,t}]
\end{aligned}
\right.
\end{equation}

Similar to (\ref{eq:motionCP})-(\ref{eq:motionCTRV}), motion functions $g^d_\mathbf{s}$ can be defined as
\begin{align}
g^\mathrm{CP}_\mathbf{s}(\mathbf{s}^\mathrm{CP}_{i,t}) & = \mathbf{s}^\mathrm{CP}_{i,t} \\
g^\mathrm{CV}_\mathbf{s}(\mathbf{s}^\mathrm{CV}_{i,t}) & = \begin{bmatrix}
                                      x^\mathrm{CV}_{i,t} + v^\mathrm{CV}_{i,t}{\rm cos}(\theta^\mathrm{CV}_{i,t}) \Delta T \\
                                      y^\mathrm{CV}_{i,t} \\
                                      z^\mathrm{CV}_{i,t} + v^\mathrm{CV}_{i,t}{\rm sin}(\theta^\mathrm{CV}_{i,t}) \Delta T \\ 
                                      \theta^\mathrm{CV}_{i,t}
                                      \end{bmatrix}
\end{align}
\begin{equation}
\begin{aligned}
& g^\mathrm{CTRV}_\mathbf{s}(\mathbf{s}^\mathrm{CTRV}_{i,t}) = \\
                                      &\begin{bmatrix}
                                      x^\mathrm{CTRV}_{i,t} + v^\mathrm{CTRV}_{i,t}{\rm cos}(\theta^\mathrm{CTRV}_{i,t} + \frac{\omega^\mathrm{CTRV}_{i,t} \Delta T}{2}) \Delta T \\
                                      y^\mathrm{CTRV}_{i,t} \\
                                      z^\mathrm{CTRV}_{i,t} + v^\mathrm{CTRV}_{i,t}{\rm sin}(\theta^\mathrm{CTRV}_{i,t} + \frac{\omega^\mathrm{CTRV}_{i,t} \Delta T}{2}) \Delta T \\ 
                                      \theta^\mathrm{CTRV}_{i,t} + \omega^\mathrm{CTRV}_{i,t} \Delta T
                                      \end{bmatrix}\\
\end{aligned}
\end{equation}

Therefore, for each motion model $d \in \mathcal{M}$, the system error for object $i$ between consecutive frames is computed as
\begin{equation}
\mathbf{e}^{\mathrm{sys},d}_{i,t,t+1} = (\mathbf{o}^d_{i,t+1})^T - g^d_\mathbf{s}(\mathbf{s}^d_{i,t})
\end{equation}

Specially for the CV and CTRV models, we further attribute the constant motion error:
\begin{equation}
\mathbf{e}^{\mathrm{cst},d}_{i,t,t+1} = \mathbf{v}^d_{i,t+1} - \mathbf{v}^d_{i,t} \ ,\ d \in \{\mathrm{CV}, \mathrm{CTRV}\} \\
\end{equation}

Finally, our objective function for optimization takes into account the model weights computed in the previous module, which can be expressed as
\begin{equation}
\mathrm{min} \sum_{t} ( \left \| \mathbf{e}^{\mathrm{odo}}_{t,t+1} \right \|^2  + \sum_{j} \left \| \mathbf{e}^{\mathrm{repr}}_{j,t} \right \|^2 + \sum_{i} \sum_{d \in \mathcal{M}} w^d_{i,t} e^d_{i,t} )
\end{equation}
where
\begin{align*}
&e^\mathrm{CP}_{i,t} = \left \| \mathbf{e}^\mathrm{repr, CP}_{i,t} \right \|^2 + \left \| \mathbf{e}^\mathrm{sys, CP}_{i,t,t+1} \right \|^2 \\
&e^\mathrm{CV}_{i,t} = \left \| \mathbf{e}^\mathrm{repr, CV}_{i,t} \right \|^2 + \left \| \mathbf{e}^\mathrm{sys, CV}_{i,t,t+1} \right \|^2 + \left \| \mathbf{e}^{\mathrm{cst, CV}}_{i,t,t+1} \right \|^2 \\
&e^\mathrm{CTRV}_{i,t} = \left \| \mathbf{e}^\mathrm{repr, CTRV}_{i,t} \right \|^2 + \left \| \mathbf{e}^\mathrm{sys, CTRV}_{i,t,t+1} \right \|^2 + \left \| \mathbf{e}^{\mathrm{cst, CTRV}}_{i,t,t+1} \right \|^2
\end{align*}

\subsection{Baseline at Methodology Level 2}

To compare different methodology levels, we implement three lower-level baselines based on visual sensors in addition to the Methodology Level 3. Specifically, the Methodology Level 2 is also based on the coupled SLAMMOT paradigm. However, current works \cite{CubeSLAM, DynaSLAM-2, VDO-SLAM, MOTSLAM} only employ a single constant velocity model to describe the object motions. Methods proposed in different studies differ in the detailed structure. For fairness of comparative study, we remove the CP and CTRV models from our IMM-based modules, serving as the implementation of the baseline at the Methodology Level 2.

\begin{figure}[!ht]
\centering
\includegraphics[width=0.43\textwidth]{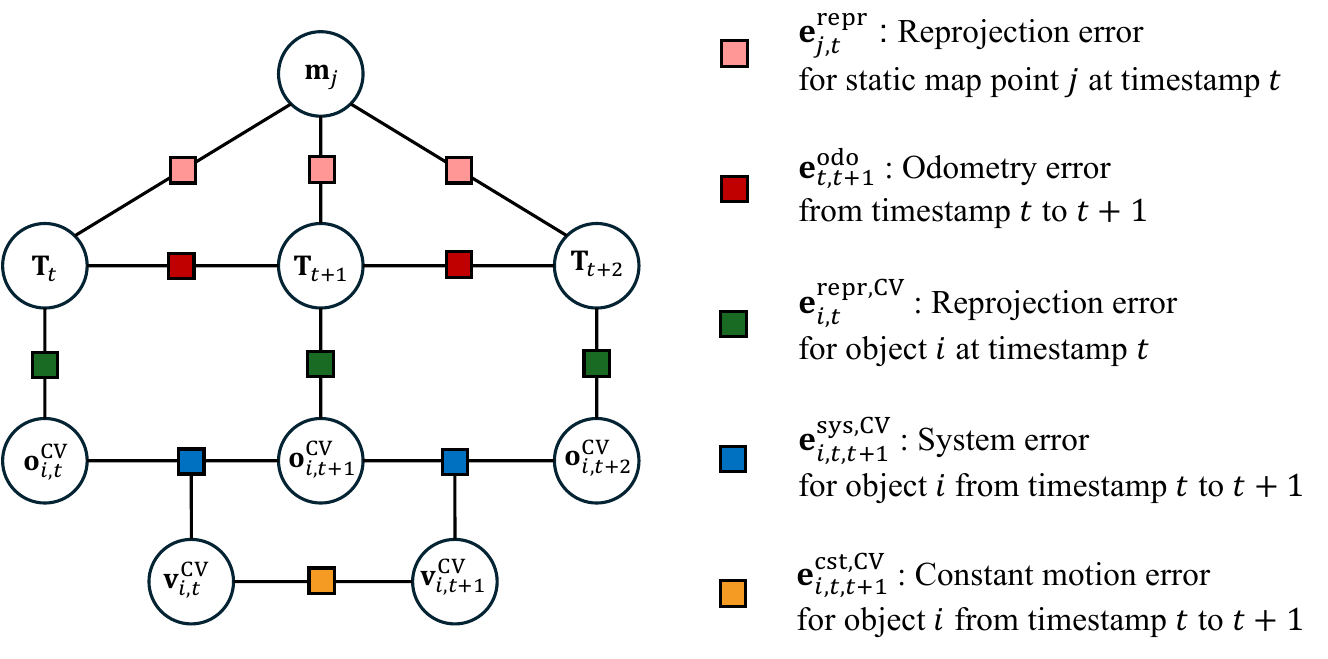}
\caption{Factor graph representation for Baseline at the Methodology Level 2.}
\label{graph-lv2}
\end{figure}

This baseline is consistent with the Methodology Level 3 in the SLAM and MOT
modules, yet is simplified in the IMM-based modules. On the one hand, in the IMM-Based Object State Estimation module, the Methodology Level 2 only includes a single step \textit{Distributed estimation}, and in (\ref{eq:EKF1})-(\ref{eq:EKF5}), we have $d = \mathrm{CV}$ only. On the other hand, in the Graph Optimization module, the factor graph of the Methodology Level 2 does not contain nodes and edges related to CP and CTRV models, as shown in Fig. \ref{graph-lv2}. The corresponding objective function is defined as (\ref{eq:cost-lv2}), without considering model weights.
\begin{equation}
\label{eq:cost-lv2}
\mathrm{min} \sum_{t} ( \left \| \mathbf{e}^{\mathrm{odo}}_{t,t+1} \right \|^2  + \sum_{j} \left \| \mathbf{e}^{\mathrm{repr}}_{j,t} \right \|^2 + \sum_{i} e^\mathrm{CV}_{i,t} )
\end{equation}

In our presented works, we would like to advocate Methodology Level 3 for visual SLAMMOT, instead of any specific implementation of visual SLAMMOT. Each of the four Methodology Levels does not depend on any composing module that uses certain specific method, but can take advantage of whatever methods to realize composing modules. For example, the four Methodology Levels can depend on a SLAM module using whatever method reviewed in Section II-A, on an image segmentation module using whatever method reviewed in Section II-B. In the context of visual SLAM, we would like to advocate Methodology Level 3 for its methodology advantage over Methodology Level 0 to 2. For fairness of comparative study, we intentionally instantiate the four methodology levels in the way that they share the same composing modules, so that their performance comparison will not be biased by any ad hoc composing module.

\section{Experiments}
\label{sec:exp}

In this section, we provide comprehensive experimental results to demonstrate the performance of our method in ego localization and multi-object tracking. We conduct experiments on other low-level baselines as well under the same conditions. Specifically, we employ ORB-SLAM2 as the Methodology Level 0 baseline method, without performing semantic analysis or special processing on the feature points in the environment. For the Methodology Level 1, we employ SpatialEmbedding\cite{SpatialEmbedding} as the segmentation module and utilize MonoDLE\cite{MonoDLE} and MoMA-M3T\cite{MoMA-M3T} for 3D object detection and tracking. For the Methodology Level 2, we implement a factor graph to tightly couple SLAM with MOT, while still using a single constant velocity motion model. For the Methodology Level 3, we incorporate IMM into the factor graph, which has been detailed in the previous section.

The experiments are conducted on the KITTI Tracking dataset, which contains 21 labeled sequences. We follow \cite{QD-3DT, MoMA-M3T} to divide the dataset into a training set (13 sequences) and a validation set (8 sequences). Several sequences in the validation set are too simple to demonstrate performances of the methods, thus we ultimately retain four representative sequences for experimentation. Sequence 0001 mostly contains stationary vehicles with a few moving ones. Sequence 0004 includes a significant number of oncoming vehicles, with the ego vehicle at a considerable distance. Sequence 0011 simultaneously contains a substantial number of stationary and moving vehicles, and Sequence 0018 represents a highway scene with a large number of moving vehicles. In each of these sequences, we further select a segment with complex variations in motion patterns to compare the performance of different methods.

It is worth noting that visual odometry based on feature points exhibits certain randomness during operation. So we carry out 50 Monte Carlo trials for each method in each scenario and evaluate the average results of these experiments.

\subsection{Ego Localization}

\begin{figure}[!ht]
\centering
\includegraphics[width=0.4\textwidth]{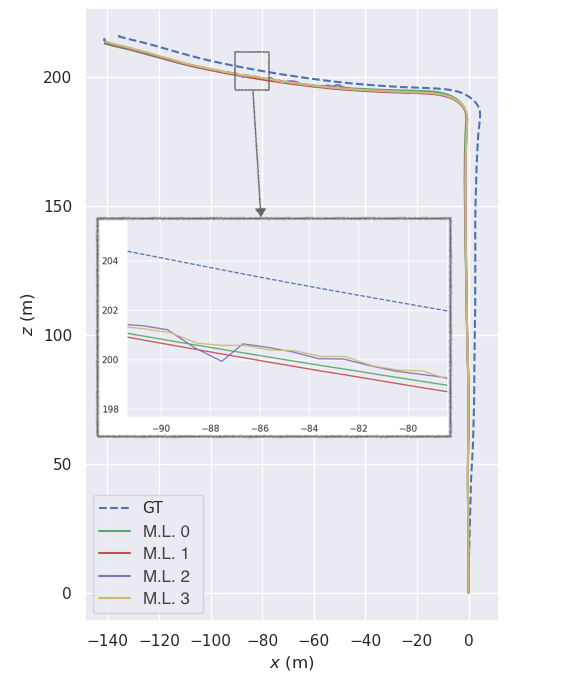}
\caption{Trajectory estimation on KITTI Tracking sequence 0001.}
\label{001-traj}
\end{figure}

In the ego localization experiments, we evaluate the performance of different methodology levels in estimating the ego vehicle's state. The evaluation metrics include Absolute Pose Error (APE) and Relative Pose Error (RPE). APE reflects the accuracy of the estimated global trajectory, while RPE reflects that of the estimated state changes over minimal time intervals.

In the implementation of the four methodological levels, the ego poses are always updated through graph optimization. The Methodology Level 0 directly employs ORB-SLAM2, where the factor graph includes ego poses and map points, regardless of whether the map points are static or dynamic. The factor graph in the Methodology Level 1 has the same types of nodes as in the Methodology Level 0, yet semantic information is used to sift out dynamic features and the map points are all static. The structure of the factor graph in the Methodology Level 2 is shown in Fig. \ref{graph-lv2}, where object information is jointly optimized to assist in ego localization. The factor graph in the Methodology Level 3 is shown in Fig. \ref{graph}, where the states of objects under different motion models are incorporated into the optimization process.

\begin{table}[!htp]
\centering
\caption{Ego Motion Comparison on KITTI Tracking Dataset}
\label{vo_seqs}
\begin{tabular}{ccccccc}
\toprule
\multicolumn{2}{c}{\textbf{Sequences}} & 0001            & 0004            & 0011            & 0018            & Mean            \\
\midrule
\multirow{2}{*}{\begin{tabular}[c]{@{}c@{}}\textbf{M.L. 0}\end{tabular}}   & APE (m)   & 5.56            & 1.39            & 2.36            & 1.48            & 2.70            \\
                           & RPE (m/f)   & 0.0337          & \textbf{0.0581} & 0.0266          & 0.0369          & 0.0389          \\
\midrule
\multirow{2}{*}{\begin{tabular}[c]{@{}c@{}}\textbf{M.L. 1}\end{tabular}}   & APE (m)   & 5.57            & 1.49            & 2.32            & 1.40            & 2.69            \\
                           & RPE (m/f)   & \textbf{0.0328} & 0.0582          & \textbf{0.0264} & \textbf{0.0357} & \textbf{0.0383} \\
\midrule
\multirow{2}{*}{\begin{tabular}[c]{@{}c@{}}\textbf{M.L. 2}\end{tabular}}   & APE (m)   & 5.52            & 5.08            & 2.29            & 1.34            & 3.56            \\
                           & RPE (m/f)   & 0.0485          & 0.0680          & 0.0454          & 0.0499          & 0.0529          \\
\midrule
\multirow{2}{*}{\begin{tabular}[c]{@{}c@{}}\textbf{M.L. 3}\end{tabular}}   & APE (m)   & \textbf{5.47}   & \textbf{1.37}   & \textbf{2.25}   & \textbf{1.32}   & \textbf{2.60}   \\
                           & RPE (m/f)   & 0.0337          & 0.0624          & 0.0299          & 0.0555          & 0.0454          \\
\bottomrule 
\end{tabular}
\end{table}

Table \ref{vo_seqs} presents the ego localization results of the four methodology levels across the four KITTI Tracking sequences, where \textbf{M.L. 0}, \textbf{M.L. 1}, \textbf{M.L. 2} and \textbf{M.L. 3} represents the Methodology Level 0, the Methodology Level 1, the Methodology Level 2, and the Methodology Level 3 respectively. For the global localization performance, the Methodology Level 3 achieves lower APE than the other methodology levels in various scenarios. Besides, we analyse performance over motion-pattern-transition periods, noting that such kinds of moments are especially critical in real traffics. As shown in Table \ref{vo_segs}, the Methodology Level 3 demonstrates even more evident advantage over the other methodology levels, reflecting its merit in handling the inherent non-deterministic nature of moving objects. 

\begin{table}[!htp]
\centering
\caption{Ego Motion Comparison during motion-pattern-transition}
\label{vo_segs}
\begin{tabular}{@{}cccccc@{}}
\toprule
\multicolumn{2}{c}{\textbf{Segments}}        & \begin{tabular}[c]{@{}c@{}}0001-\\ {[}365, 420{]}\end{tabular} & \begin{tabular}[c]{@{}c@{}}0004-\\ {[}0, 30{]}\end{tabular} & \begin{tabular}[c]{@{}c@{}}0011-\\ {[}130, 190{]}\end{tabular} & \begin{tabular}[c]{@{}c@{}}0018-\\ {[}295, 320{]}\end{tabular} \\ \midrule
\multirow{2}{*}{\textbf{M.L. 0}} & APE (m)   & 14.07                                                          & 0.17                                                         & 3.02                                                           & 4.29                                                           \\
                                 & RPE (m/f) & \textbf{0.0122}                                                & 0.0288                                                       & 0.0358                                                         & \textbf{0.0106}                                                \\ \midrule
\multirow{2}{*}{\textbf{M.L. 1}} & APE (m)   & 14.26                                                          & 0.17                                                         & 3.80                                                           & 4.25                                                           \\
                                 & RPE (m/f) & 0.0123                                                         & 0.0291                                                       & \textbf{0.0346}                                                & 0.0122                                                         \\ \midrule
\multirow{2}{*}{\textbf{M.L. 2}} & APE (m)   & 22.82                                                          & 0.21                                                         & 2.98                                                           & 5.78                                                           \\
                                 & RPE (m/f) & 0.0801                                                         & 0.0332                                                       & 0.0521                                                         & 0.0455                                                         \\ \midrule
\multirow{2}{*}{\textbf{M.L. 3}} & APE (m)   & \textbf{13.92}                                                 & \textbf{0.12}                                                & \textbf{2.90}                                                  & \textbf{3.56}                                                  \\
                                 & RPE (m/f) & 0.0156                                                         & \textbf{0.0285}                                              & 0.0357                                                         & 0.0358                                                         \\  \bottomrule
\end{tabular}
\end{table}

We also observe that trajectories estimated by the Methodology Level 2 and the Methodology Level 3 exhibit larger local fluctuations (as illustrated in Fig. \ref{001-traj}) and hence have larger RPE. Roughly speaking, the phenomenon is due to comparatively high fluctuations that usually exist in visual measurements of moving objects
\footnote{In contrast, LiDAR range measurements of moving objects tend to be more stable and accurate.}. Despite such inherent limitation of vision systems, the Methodology Level 3 can considerably alleviate local fluctuations, thanks to adaptiveness and flexibility brought by consideration of multiple motion models. Besides, it is worth noting that performance in APE is more important than performance in RPE, because it is the former that essentially reflects the SLAMMOT's long-term localization ability, whereas local fluctuations would after all be further alleviated via certain kind of filtering in practice.

As we can see, even suppose estimation results of moving objects are not cared in practice, visual SLAMMOT in the spirit of the Methodology Level 3 can still be beneficial in terms of providing an advantageous ego localization function (or virtual odometry).

\subsection{Multi-Object Tracking}

In multi-object tracking experiments, we evaluate the performance of different levels of methodologies in estimating the states of dynamic objects. Our primary concern is the MOTP metric, which reflects the discrepancy between the estimated and true positions of objects. CLEAR MOT\cite{CLEARMOT} has introduced other common MOT metrics, such as MOTA and ID switch. However, the results obtained for these metrics are solely dependent on the efficacy of the 3D MOT models themselves. Since the same models (MonoDLE and MoMA-M3T) have been used for prior 3D MOT inference across methodologies from Level 1 to Level 3, they are consistent across these levels and are not presented and compared in this paper.

In our implementation, the object state estimation in the Methodology Level 1 primarily derives from the output of the MOT module. This output is then used to estimate the position and yaw rate of objects in the world coordinate system, utilizing localization information provided by the SLAM module after filtering out dynamic objects. For the Methodology Level 2, the output from the MOT module is jointly optimized with SLAM localization and mapping results to achieve a tightly coupled SLAMMOT framework. However, in the Methodology Level 2, both object state estimation and graph optimization use a single CV model to describe object states. In contrast, the Methodology Level 3 incorporates three motion models (CP, CV, CTRV) into the object state estimation and graph optimization processes.

\begin{table}[!htp]
\centering
\caption{Object Motion Comparison on KITTI Tracking Dataset}
\label{mot_seqs}
\begin{tabular}{@{}ccccccc@{}}
\toprule
\multicolumn{2}{c}{\textbf{Sequences}}                                            & 0001          & 0004          & 0011          & 0018          & Mean          \\
\midrule
\begin{tabular}[c]{@{}c@{}}\textbf{M.L. 1}\end{tabular}   & MOTP (m) & 2.33          & 1.83          & 4.74          & 1.89          & 2.70          \\
\midrule
\begin{tabular}[c]{@{}c@{}}\textbf{M.L. 2}\end{tabular}   & MOTP (m) & 2.27          & 2.02          & 4.63          & 1.90          & 2.71          \\
\midrule
\begin{tabular}[c]{@{}c@{}}\textbf{M.L. 3}\end{tabular} & MOTP (m) & \textbf{2.08} & \textbf{1.75} & \textbf{4.53} & \textbf{1.88} & \textbf{2.56} \\
\bottomrule
\end{tabular}
\end{table}

Table \ref{mot_seqs} presents the MOT performances of three levels of methodologies across four different sequences. Based on general results, the Methodology Level 3 consistently achieves the lowest MOTP, indicating that our approach can achieve accurate object state estimation while accurately estimating ego vehicle's trajectory, thereby realizing mutual benefits of SLAM and MOT. Besides, Table \ref{mot_segs} presents a comparison of different methods in segments with motion pattern changes. The advantages of the Methodology Level 3 in MOT become more significant, highlighting the effectiveness of this methodology in complex and dynamic outdoor scenarios.

\begin{table}[!thp]
\centering
\caption{Object Motion Comparison during motion-pattern-transition}
\label{mot_segs}
\begin{tabular}{@{}cccccc@{}}
\toprule
\multicolumn{2}{c}{\textbf{Segments}}        & \begin{tabular}[c]{@{}c@{}}0001-\\ {[}365, 420{]}\end{tabular} & \begin{tabular}[c]{@{}c@{}}0004-\\ {[}0, 30{]}\end{tabular} & \begin{tabular}[c]{@{}c@{}}0011-\\ {[}130, 190{]}\end{tabular} & \begin{tabular}[c]{@{}c@{}}0018-\\ {[}295, 320{]}\end{tabular} \\ 
\midrule
\begin{tabular}[c]{@{}c@{}}\textbf{M.L. 1}\end{tabular}   & MOTP (m) & 2.09          & 1.50          & 6.67          & \textbf{1.02}          \\
\midrule
\begin{tabular}[c]{@{}c@{}}\textbf{M.L. 2}\end{tabular}   & MOTP (m) & 2.03          & 1.13          & 6.57          & 1.03          \\
\midrule
\begin{tabular}[c]{@{}c@{}}\textbf{M.L. 3}\end{tabular} & MOTP (m) & \textbf{1.41} & \textbf{1.10} & \textbf{6.42} & \textbf{1.02} \\
\bottomrule
\end{tabular}
\end{table}

\subsection{Computation Overhead}

In our approach, deep learning models are utilized for image segmentation and 3D multi-object tracking. These models can be accelerated by GPUs and generally meet real-time operation requirements when deployed on in-vehicle hardware. Our method is implemented in C++ and the experiments are conducted on a 2.3GHz i7-11800H CPU. For a single frame, the visual odometry takes approximately 3ms, while the coupled SLAMMOT framework based on EKF and graph optimization consumes about 40ms. Thus, our method meets the real-time requirements in practical application scenarios. 

\subsection{Special Observations of Visual SLAMMOT}

In the presented work, we have some special observations of visual SLAMMOT in outdoor traffic environments, which imply that vision-based systems are indeed different from LiDAR-based systems (see our team's previous work\cite{IMM-SLAMMOT}), in terms of comparison among the four methodology levels.

For LiDAR-based systems operating in dynamic environments, the Methodology Level 0 tends to have inferior performance (in terms of ego localization) than the other methodology levels. However, for vision-based systems, the Methodology Level 0 does not have such tendency of being inferior and may even outperform the Methodology Level 1 and the Methodology Level 2. In other words, the Methodology Level 0 is not so ``bad'' in outdoor traffic environments, though it rigidly treats all objects as stationary in SLAM. Besides, for vision-based systems, the Methodology Level 2 is especially susceptible to measurement uncertainties.

Some heuristic explanations hover over above observations. A LiDAR is indeed accurate but has a limited perception range (usually dozens of meters), so distant environment parts beyond the perception range are invisible to the LiDAR. In contrast, a camera has no limitation in the perception range and has broader perception of static environment parts: it can even extract static features far away on cloud. During operation in outdoor traffic environments, static visual features are usually abundant and tend to maintain performance of the Methodology Level 0.

Compared with LiDAR range measurements of moving objects, visual measurements usually have considerably higher fluctuations, especially for distant moving objects. Generally speaking, tracking based on a single motion model (conventionally a high-order single motion model) tends to converge slowly\cite{Li2022FARET} \cite{FARET}. Visual measurements with high fluctuations worsen the slow convergence problem further, causing severe fluctuations in estimation results, especially during motion-pattern-transition periods. This may account for observations that for vision-based systems, the Methodology Level 2 is susceptible to measurement uncertainties and does not demonstrate similar advantage over the Methodology Level 0 and the Methodology Level 1 as for LiDAR-based systems.

Thanks to adaptiveness and flexibility brought by consideration of multiple motion models to visual SLAMMOT, the Methodology Level 3  still demonstrates its advantages over the other methodology levels in terms of both ego localization and multi-object tracking, like how  the Methodology Level 3 benefit LiDAR-based SLAMMOT\cite{IMM-SLAMMOT}.

\section{Conclusion and Future Work}
\label{sec:conclusion}

In this paper, we have instantiated the Methodology Level 3 as visual SLAMMOT, which tightly couples SLAM and MOT, jointly optimizing the ego poses, static map, and dynamic object states in a novel bundle adjustment framework. Experimental results demonstrate that the Methodology Level 3 achieves improvements in localization and object tracking performances compared to the lower-level baselines. This advantage is evident in typical outdoor dynamic scenarios, and is significant when surrounding objects exhibit varying motion patterns. This indicates that Methodology Level 3 realizes mutual benefits between SLAM and MOT, and the consideration of multiple motion models enhances the description of dynamic objects. In real-world applications, Methodology Level 3 can be applied with various sensors and can easily be combined with the state-of-the-art MOT models to achieve further improvements. Overall, the Methodology Level 3 can perform in real-time the localization, mapping, and object tracking tasks in a unified framework, providing rich self and environmental information for intelligent vehicles.

Based on the current framework, further work could attempt to incorporate more perception information to improve the accuracy of estimates. For instance, the image segmentation model has the potential to provide dense 2D object information, and the bounding boxes obtained from the 3D detection model could be considered for optimization. In addition, more efforts could focus on reducing the fluctuation in trajectory estimation, enabling smoother ego motion prediction while maintaining localization accuracy.

For the moment, we still do not have a systematic mechanism to perform on-line estimation of state uncertainty for Methodology Level 3 (in fact, not for the other Methodology Levels either). However, for practical applications, to systematically know uncertainty of the state estimate is as important as to estimate the state value itself, though empirical setting may be an expedient practice. To research on such a systematic mechanism can be a valuable direction for further extensions.

\addtolength{\textheight}{-2cm}   


\bibliographystyle{ieeetr}
\bibliography{references}

\end{document}